# Query Efficient Posterior Estimation in Scientific Experiments via Bayesian Active Learning

**Kirthevasan Kandasamy, Jeff Schneider, Barnabás Póczos**
Carnegie Mellon University, Pittsburgh PA
{kandasamy, schneide, bapoczos}@cs.cmu.edu

**Abstract**

A common problem in disciplines of applied Statistics research such as Astrostatistics is of estimating the posterior distribution of relevant parameters. Typically, the likelihoods for such models are computed via expensive experiments such as cosmological simulations of the universe. An urgent challenge in these research domains is to develop methods that can estimate the posterior with few likelihood evaluations.

In this paper, we study active posterior estimation in a Bayesian setting when the likelihood is expensive to evaluate. Existing techniques for posterior estimation are based on generating samples representative of the posterior. Such methods do not consider efficiency in terms of likelihood evaluations. In order to be query efficient we treat posterior estimation in an active regression framework. We propose two myopic query strategies to choose where to evaluate the likelihood and implement them using Gaussian processes. Via experiments on a series of synthetic and real examples we demonstrate that our approach is significantly more query efficient than existing techniques and other heuristics for posterior estimation.

# 1 Introduction

Computing the posterior distribution of parameters given observations is a central problem in Bayesian statistics. We use the posterior distribution to make inferences about likely parameter values and estimate functionals we are interested in. For simple parametric models with conjugate priors we may obtain the posterior in analytic form. In more complex models where the posterior is analytically intractable, we have to resort to approximation techniques. In some cases, we only have access to a black box which computes the likelihood for a given value of the parameters.

Our goal is an efficient way to estimate posterior densities when calls to this black box are expensive. This work is motivated by applications in computational physics and cosmology. Several cosmological phenomena are characterised by the cosmological parameters (e.g. Hubble constant, dark energy fraction). Given observations, we wish to make inferences about the parameters. Physicists have developed simulation-based probability models of the Universe which can be used to compute the likelihood of cosmological parameters for a given observation. Figure 1 shows different scenarios to estimate/compute the likelihood. Many problems in scientific computing have a similar flavour. Expensive simulators in molecular mechanics, computational biology and neuroscience are used to model many scientific processes. Hence this work finds relevance in these fields as well.

**Related Work**

Practitioners have conventionally used sampling schemes [16] to approximate the posterior distributions. Rejection sampling and various MCMC methods are common choices. The advantage of MCMC approaches is their theoretical guarantees with large sample sets [24] and thus they are a good choice when likelihood evaluations are cheap. However, none of them is intended to be query efficient when evaluations are expensive. Some methods spend most of their computation evaluating point likelihoods and then discard the likelihood values after doing an acceptance test. This gives insight into the potential gains possible by retaining those likelihoods for use in regression. Despite such deficiencies, MCMC remains one of the most popular techniques for posterior estimation in experimental physics [6, 13, 21] and the other fields [14].

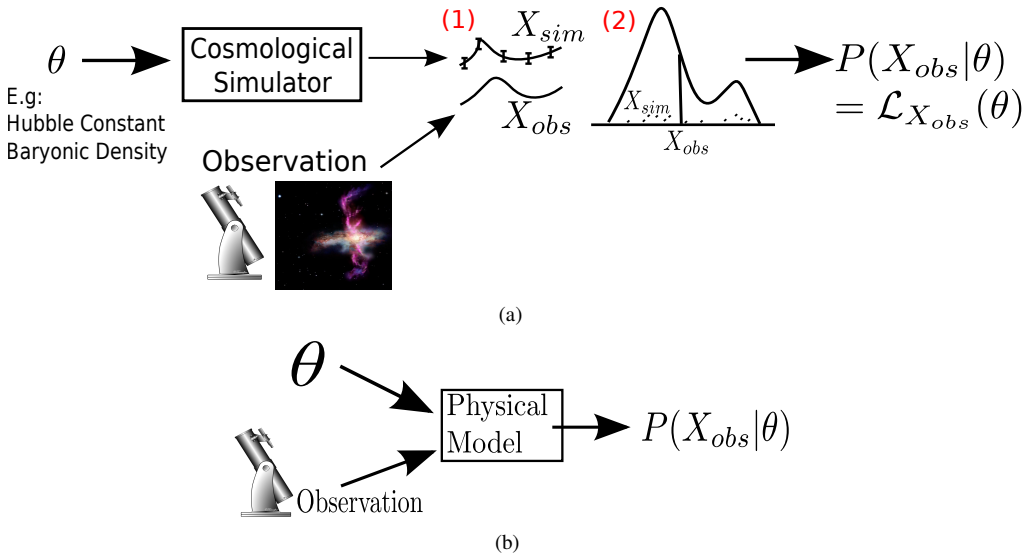

Figure 1: **Illustrations of Cosmological Experiments.** (a): Given a set of values for the parameters $\theta$ the oracle produces several simulations $\mathbf{X_{sim}}$. The likelihood $P(\mathbf{X_{obs}}|\theta)$ can then be estimated, say, by (1) comparing $\mathbf{X_{sim}}$ to the $\mathbf{X_{obs}}$ via a statistical test or (2) constructing a density estimate using $\mathbf{X_{sim}}$ and then evaluating this estimate at $\mathbf{X_{obs}}$. (b): The oracle directly computes the likelihood using a physical model of the universe.

Approximate Bayesian computation (ABC) [17, 18] is a method of last resort for estimating posteriors when a likelihood can not be computed. Unfortunately, it still requires the generation of simulated data, which is expensive in our setup, and it does not address efficient selection of parameter values to be tested at all. Nested Sampling [27] is a technique commonly used is Astrostatistics. Kernel Bayes' Rule [7] is a non-parametric method of computing a posterior based on the embedding of probabilities in an RKHS. Both these methods require sampling from a distribution and do not address the question of which samples to choose if generating them is expensive. The work in Bryan et al. [3] actively learns level sets of an expensive function and derives confidence sets from the results. Gotovos et al. [8] also actively learn level sets via a classification approach. Our work is more general since we estimate the entire posterior.

Our methods draw inspiration from Gaussian Process (GP) based active learning methods such as Bayesian optimisation (BO) [19], Bayesian quadrature (BQ) [20], active GP Regression (AGPR) [25] and several others [10–12, 15, 28]. These methods have a common modus operandi to determining the experiment $\theta_t$ at time step $t$: construct a utility function $u_t$ based on the posterior GP conditioned on the queries so far; then maximise $u_t$ to determine $\theta_t$. $u_t(\theta)$ captures the value of performing an experiment at point $\theta$. Maximising the typically multimodal $u_t$ is itself a hard problem. Further, inference in GPs can also be quite expensive, especially when we have several queries. However, it is generally assumed that querying the function is more costly than the effort to determine the next experiment [2, 28]. The key difference in such methods is essentially in the specification of $u_t$ to determine the next experiment. In our work, we adopt this strategy. We present two utility functions for active posterior estimation.

**Our contribution** is to propose a query efficient method for estimating posterior densities when the likelihood is expensive to evaluate. We adopt a Bayesian active regression approach on the log likelihood using the samples it has already computed. We refer to this approach as Bayesian Active Posterior Estimation (BAPE). We propose two myopic query strategies on the uncertainty regression model for sample selection. Our implementation uses Gaussian processes [23] and we demonstrate the efficacy of the methods on multiple synthetic and real experiments.

The remainder of this manuscript is organised as follows. We begin with a brief review of GPs in Section 2.1. We present our methods in Sections 2.2, 2.3, 2.4 and compare them against non-active strategies and GP based active learning methods in Section 2.5. In Section 3 we discuss alternatives for empirical evaluation and in Section 4 we present experimental results on synthetic and real problems.

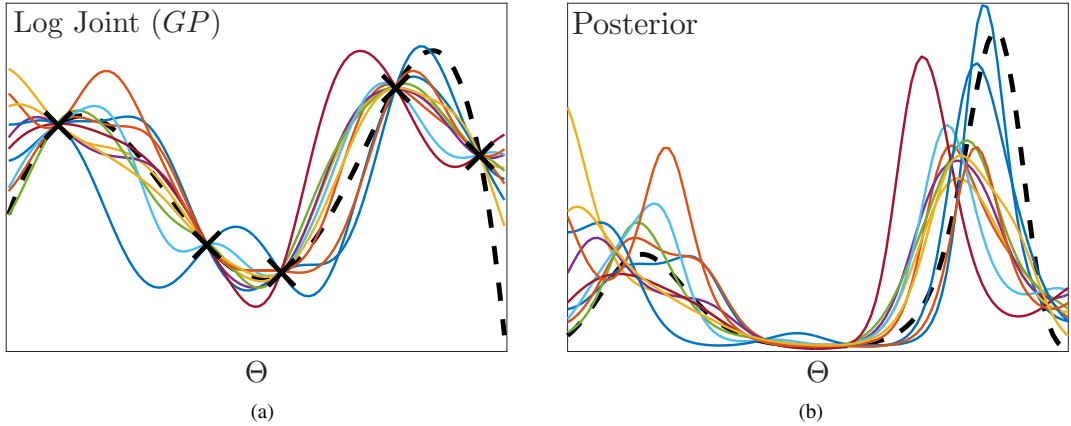

Figure 2: (a) depicts the uncertainty for the log joint probability via samples $g$ drawn from the GP. (b) illustrates the induced uncertainty model $F_{\theta|\mathbf{X_{obs}}}$ for the posterior via the exponentiated and normalised samples $f = \exp g / \int \exp g$.

## 2 Bayesian Active Posterior Estimation

### 2.1 Gaussian Processes

A GP over a space $\Theta$ is a random process describing functions from $\Theta$ to $\mathbb{R}$. It is completely characterised by its mean function $\mu : \Theta \to \mathbb{R}$ and covariance kernel $k : \Theta \times \Theta \to \mathbb{R}$. The function values $f(\theta_1), \ldots, f(\theta_n)$ at any finite set of $n$ points $\{\theta_1, \ldots, \theta_n\} \subset \Theta$ are distributed jointly Gaussian $\mathcal{N}(\boldsymbol{\mu}, \mathbf{K})$. Here $\boldsymbol{\mu} \in \mathbb{R}^n$, $\boldsymbol{\mu}_i = \mu(\theta_i)$ and $\mathbf{K} \in \mathbb{R}^{n \times n}$ where $\mathrm{K}_{ij} = k(\theta_i, \theta_j)$.

Typically, a GP is specified by its prior mean $\mu_0$ and kernel $k_0$. Given observations $\mathbf{y} = (y_1, \ldots, y_n)^\top \in \mathbb{R}^n$ at points $\boldsymbol{\theta} = \{\theta_1, \ldots, \theta_n\}$ the posterior is also a GP with mean $\mu'$ and covariance $k'$ given by,

$$\mu'(\theta) = \mu_0(x) + k_0(\boldsymbol{\theta}, \theta)^\top \mathbf{K}_0^{-1}(\mathbf{y} - \mu_0(\boldsymbol{\theta})),$$
$$k'(\theta, \theta') = k_0(\theta, \theta') - k_0(\boldsymbol{\theta}, \theta)^\top \mathbf{K}_0^{-1} k_0(\boldsymbol{\theta}, \theta').$$

The posterior variance at any $\theta \in \Theta$ is given by $k'(\theta, \theta)$. Here, we have overloaded notation to denote, $\mu_0(\boldsymbol{\theta}) = (\mu_0(\theta_1), \ldots, \mu_0(\theta_n))^\top \in \mathbb{R}^n$, $k_0(\boldsymbol{\theta}, \theta) = (k_0(\theta_1, \theta), \ldots, k_0(\theta_n, \theta))^\top \in \mathbb{R}^n$, and $\mathbf{K}_0 \in \mathbb{R}^{n \times n}$ whose $(i,j)^{\text{th}}$ element is $k_0(\theta_i, \theta_j)$. A popular choice for the covariance kernel is the squared exponential kernel $k_{\sigma_f, h}(\theta, \theta') = \sigma_f^2 \exp(-\|\theta - \theta'\|^2/(2h^2))$. When the observations of $f$ are noisy, it is common to use a noise term in the prior kernel $k_0 = k_{\sigma, h} + \sigma_n^2 \delta(\theta - \theta')$ where $\delta$ is the dirac-delta function. This corresponds to assuming white Gaussian noise with variance $\sigma_n^2$. For a more detailed review of GPs we recommend Rasmussen and Williams [23]. Following other GP based active learning methods [12, 28], we use GPs primarily as an uncertainty model for an unknown function.

### 2.2 Active Posterior Estimation

**Problem Setting:** We formally define our posterior distribution estimation problem in a Bayesian framework. We have a bounded continuous parameter space $\Theta$ for the unknown parameters (e.g. cosmological constants). Let $\mathbf{X_{obs}}$ denote our observations (e.g. signals from telescopes). For each $\theta \in \Theta$ we have the ability to query an oracle for the value of the likelihood $\mathcal{L}(\theta) = P(\mathbf{X_{obs}}|\theta)$. These queries are expensive and possibly noisy. Assuming a prior $P_\theta(\theta)$ on $\Theta$, we have the posterior $P_{\theta|\mathbf{X_{obs}}}$.

$$P_{\theta|\mathbf{X_{obs}}}(\theta|\mathbf{X_{obs}}) = \frac{\mathcal{L}(\theta) P_\theta(\theta)}{\int_\Theta \mathcal{L}(\theta) P_\theta(\theta)} = \frac{\mathcal{L}(\theta) P_\theta(\theta)}{P(\mathbf{X_{obs}})} \quad (1)$$

Our goal is to obtain an estimate $\widehat{P}_{\theta|\mathbf{X_{obs}}}$ of $P_{\theta|\mathbf{X_{obs}}}$ using as few queries to the oracle.

**Algorithm 1** Bayesian Active Posterior Estimation

**Given:** Input space $\Theta$, GP prior $\mu_0$, $k_0$.
**For** $t = 1, 2, \ldots$ **do**
  1. $\theta_t = \text{argmax}_{\theta_t \in \Theta} \, u_t(\theta)$
  2. $\mathcal{L}_t \leftarrow$ Query oracle at $\theta_t$.
  3. Obtain posterior conditioned on $(\theta_i, \mathcal{L}_i P_\theta(\theta_i))_{i=1}^t$

Some smoothness assumptions on the problem are warranted to make the problem tractable. A standard in the Bayesian nonparametrics literature is to assume that the function of interest is a sample from a Gaussian Process. In what follows we shall model the log joint probability of the cosmological parameters and the observations via a GP[1]. This is keeping in line with Adams et al. [1] who use a similar prior for GP density sampling.

**Uncertainty for the posterior via uncertainty for the log joint:** Assume that we have queried the likelihood oracle at $n$ points, and for each query point $\theta_i$ the oracle provided us with $\mathcal{L}_i \approx P(\mathbf{X_{obs}}|\theta_i)$ answers. Let $A_n = \{\theta_i, \mathcal{L}_i\}_{i=1}^n$ denote the set of these query value pairs. We build our GP using $B_n = \{\theta_i, \log(\mathcal{L}_i P_\theta(\theta_i))\}_{i=1}^n$ as the input output pairs. The GP automatically induces uncertainty on the log joint probability; let us denote the distribution of $\log P(\mathbf{X_{obs}}, \theta)$ values at any $\theta \in \Theta$ by $\tilde{\mathcal{L}}(\theta)$. Moreover, if $g$ is a sample from this GP, then $f = \exp g / \int \exp g$ denotes a sample from the induced uncertainty model $F_{\theta|\mathbf{X_{obs}}}$ for the posterior $P_{\theta|\mathbf{X_{obs}}}$. A sample from $F_{\theta|\mathbf{X_{obs}}}$ is a distribution over $\Theta$. This is illustrated in Figure 2.

Finally, given any estimate $\widehat{\mathcal{P}}^{A_n}(\mathbf{X_{obs}}, \theta)$ of the log joint probability constructed using a set $A_n$ of $n$ parameter-likelihood pairs, the estimate of the posterior distribution is

$$\widehat{P}^{A_n}(\theta|\mathbf{X_{obs}}) = \frac{\exp \widehat{\mathcal{P}}^{A_n}(\mathbf{X_{obs}}, \theta)}{\int_\Theta \exp \widehat{\mathcal{P}}^{A_n}(\mathbf{X_{obs}}, \theta)}. \tag{2}$$

**Bayesian Active Posterior Estimation:** We now describe the procedure to determine the point at which we should query the likelihood. At time step $t$, we have already queried at $t-1$ points and have the set $A_{t-1}$ of query value pairs. Our goal is to select the point $\theta_t$ for the next experiment to evaluate the likelihood. We adopt a myopic strategy which picks the point that maximizes a utility function $u_t$. $u_t$ needs to capture a measure of divergence $D(\cdot \| \cdot)$ between the true and estimated distributions. A reasonable strategy would be to select $\theta_t$ to satisfy

$$\theta_t = \underset{\theta_+ \in \Theta}{\text{argmin}} \, D( P_{\theta|\mathbf{X_{obs}}} \, \| \, \widehat{P}^{A_{t-1} \cup \{(\theta_+, \mathcal{L}(\theta_+))\}} ) \tag{3}$$

where $\widehat{P}^{A_{t-1} \cup \{(\theta_+, \mathcal{L}(\theta_+))\}}$ is our estimate of the posterior using $A_{t-1} \cup \{(\theta_+, \mathcal{L}(\theta_+))\}$. Obviously, this objective is not accessible in practice, since we know neither $P_{\theta|\mathbf{X_{obs}}}$ nor $\mathcal{L}(\theta_+)$. As surrogates to this ideal objective in Equation (3), in the following subsections we propose two utility functions for determining the next point: Negative Expected Divergence (NED) and Exponentiated Variance (EV). The first, NED adopts a Bayesian decision theoretic approach akin to the expected error reduction criterion used in active learning [26]. Here, we choose the point in $\Theta$ that yields the minimum expected divergence for the next estimate over the uncertainty model. Unfortunately, as we will see in Section 2.3, the NED utility is computationally expensive. Therefore, we propose a cheaper alternative, EV. In our experiments we found that both strategies performed equally well – so EV is computationally attractive. That said, some cosmological simulations are very expensive (taking several hours to a day) so NED is justified in such situations. We present our framework for BAPE using an appropriate utility function $u_t$ in Algorithm 1.

### 2.3 Negative Expected Divergence (NED)

Equation (3) says that we should choose the point that results in the highest reduction in divergence *if we knew the likelihood and the true posterior at that point*. In NED, we choose the point with the highest expected reduction in divergence. For the next evaluation we choose the point that minimizes the expected divergence between these models and the next estimate. Precisely,

$$u_t^{\text{NED}}(\theta_+) = -\mathbb{E}_{p_+ \sim \tilde{\mathcal{L}}(\theta_+)} \mathbb{E}_{h \sim F_{\theta|\mathbf{X_{obs}}}} D(\, h \,\|\, \widehat{P}_{m+1}^{A \cup \{(\theta_+, p_+)\}} \,). \tag{4}$$

---
[1] We work in the log joint probability space since log smoothes out a function and is more conducive to be modeled as a GP. We also avoid issues such as non-negativity of $\widehat{P}_m^A(\theta|\mathbf{X_{obs}})$. Osborne et al. [20] also use a similar log-transform before applying a GP.

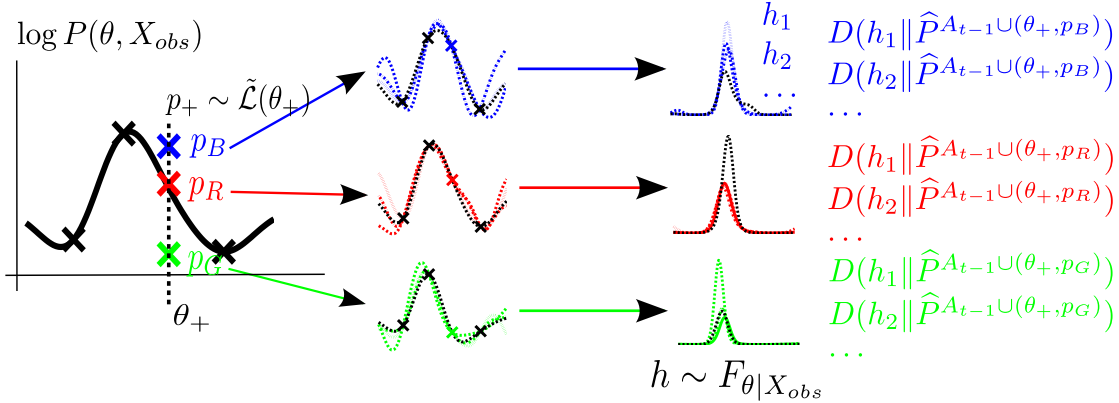

Figure 3: An illustration of the NED utility. $\theta_+$ is a candidate for the next evaluation. $p_B, p_R, p_G$ denote values for $p_+ = \log P(\theta_+, \mathbf{X_{obs}})$ sampled from the GP. We add them as hallucinated points and rebuild our GP and generate samples (second step). These samples are exponentiated and normalised (third step) and then its KL divergence with the estimate is computed.

Here $p_+ \in \mathbb{R}$ is sampled from $\tilde{\mathcal{L}}(\theta_+)$, the uncertainty of the log joing probability at $\theta_+$. The *density* $h$ is sampled from $F_{\theta|\mathbf{X_{obs}}}$, the uncertainty model of the posterior obtained by adding $(\theta_+, p_+)$ to the set of already available points $A_{t-1}$. Both $\tilde{\mathcal{L}}(\theta_+)$ and $F_{\theta|\mathbf{X_{obs}}}$ are induced from the log joint GP as explained before. $\widehat{P}_{m+1}^{A \cup \{(\theta_+, p_+)\}}$ denotes the estimate of the posterior obtained by re-training the GP with $(\theta_+, p_+)$ as the $t^{\text{th}}$ point along with the $t$ points already available. The first expectation above captures our uncertainty over $\log P(\theta_+, \mathbf{X_{obs}})$ while the second captures our remaining uncertainty over $P_{\theta|\mathbf{X_{obs}}}$ after observing $\mathcal{L}(\theta_+)$. Equation (4) says that you should minimize the expected divergence by looking one step ahead.

We have illustrated NED in Figure 3. Assume we are considering the point $\theta_+$ for the next evaluation. Our GP over the log joint probability gives us uncertainty for $\log P(\theta_+, \mathbf{X_{obs}})$ – depicted by $p_B, p_R, p_G$ in blue, green and red respectively. For $p_B$, we add $(\theta_+, p_B)$ as a hallucinated point to the $t-1$ points we already have and obtain an estimate of the posterior $\widehat{P}^{A_{t-1} \cup (\theta_+, p_B)}$. Next, we rebuild our GP using these $t$ points. We draw samples from the new GP and exponentiate and normalise them to obtain samples $h_i$ from the uncertainty model for the posterior $F_{\theta|\mathbf{X_{obs}}}$. Then we compute the divergence between $h_i$ and $\widehat{P}^{A_{t-1} \cup (\theta_+, p_B)}$. We repeat this for the blue and green points and average all the divergences. The next evaluation point will be that with the lowest expected one step ahead divergence.

The expectations in the NED utility above are computationally intractable. They can only be approximated empirically by drawing samples and require numerical integration (as Figure 3 suggests). For these reasons we propose an alternate utility function below. In our experiments we found that both EV and NED performed equally well.

## 2.4 Exponentiated Variance (EV)

A common active learning heuristic is to choose the point that you are most uncertain about for the next experiment. As before we use a GP on the log joint probability. At any given point in this GP we have an associated posterior variance of the GP. However, this variance corresponds to the uncertainty of the *log* joint probability whereas our objective is in learning the joint probability – which is a multiplicative factor away from the posterior. See Figure 4. Therefore, unlike in usual GP active learning methods [25], the variance of interest here is in the exponentiated GP. By observing that an exponentiated Gaussian is a log Normal distribution, the EV utility function is given by

$$u_t^{\text{EV}}(\theta_+) = \exp(2\mu_t(\theta_+) + \sigma_t^2(\theta_+))(\exp(\sigma_t^2(\theta_+)) - 1) \tag{5}$$

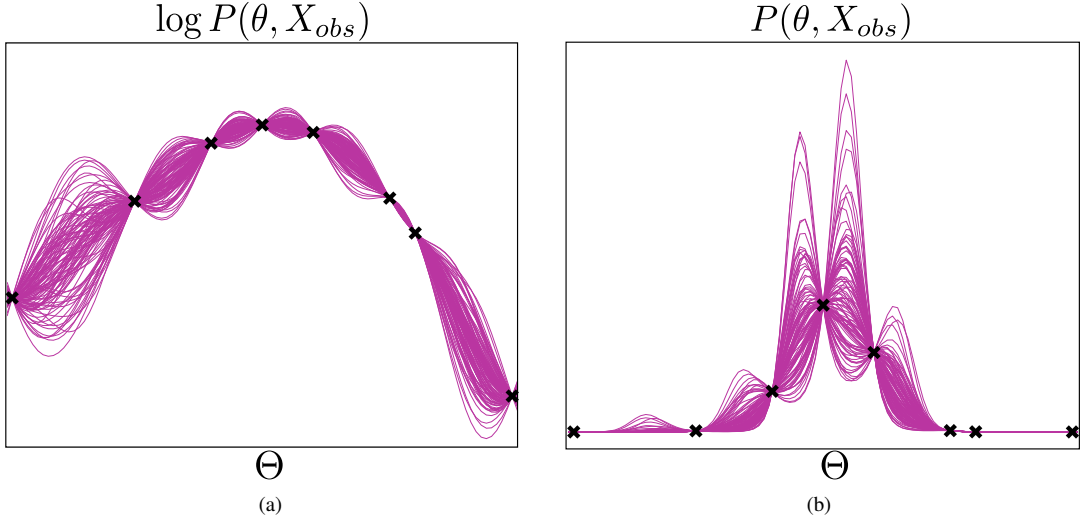

Figure 4: (a): Samples drawn from the GP in the log joint probability space. (b): The same samples after exponentiation. High variance in the low likelihood regions are squashed and low variances in the high likelihood regions are blown up. This is the key insight that inspires our methods and the EV utility in particular.

Here $\mu_t, \sigma_t^2$ are the posterior mean and variances of the GP at time $t$. The $\exp(2\mu(\theta_+))$ will squash high variances in the low likelihood regions and amplify low variances in the high likelihood regions. The expression for $u_t^{\text{EV}}(\theta_+)$ corresponds precisely to the variance of $P(\theta, \mathbf{X_{obs}})$ according to the uncertainty model induced by the GP. We choose the point *maximising* the above variance to determine the next query location.

## 2.5 Discussion

We first argue that an active, i.e. an adaptive sequential, strategy will be useful for posterior estimation. In particular, the work of Castro et al. [4] demonstrates that active learning does not perform significantly better than passive strategies to estimate a uniformly smooth function uniformly well. However, in our case we wish to learn the function well at high log probabilty regions as they predominantly determine the shape of the posterior. To illustrate this we have shown a synthetically created log joint probability and the corresponding posterior in the first column of Figure 5. The high likelihood regions largely affect the shape of the posterior since the variations in the low likelihood region are squashed after exponentiation. In the second column we queried the likelihood at uniformly spaced points and obtained estimates of the log joint probability and the posterior in green. In the third we have the same estimates in magenta, except that we used more queries at at high likelihood regions. While the green estimate for the *log joint* may be uniformly better than the magenta estimate, it is the opposite for *posterior*. This is because after exponentation, the small errors in the high log probability regions have been inflated after exponentiation for the green estimate, whereas for the magenta estimate the large errors in the low probability regions have been diminished. Hence an active strategy, which uses more queries at the high log probability regions is likely to do better than a passive strategy. NED and EV do precisely this by attaching more emphasis to the uncertainties in the high log probability regions.

Next, it is important to distinguish our objective in this work from similar active learning literature in the GP framework. In BO, the objective is to find the maximum of a function. This means that once the active learner realises that it has found the mode of a function it has less incentive to explore around as it would not improve the current maximum values. For instance, consider the log joint probability in Figure 6(a) and the joint probability in Figure 6(b). We have shown the points where we have already queried at as brown crosses and the red circles (x) and (y) show possible candidates for the next query. The shaded regions represent the uncertainty due to three standard deviations in the GP. In BO, the active learner would not be interested in (y) as, by virtue of points (5), (6) and (7) it knows that (y) is not likely to be higher than (6). On the other hand, in BAPE we are keen on (y) as knowing it with precision will significantly affect our estimate of the posterior (Fig 6(b)). In particular to know the posterior well we will need to query at the neighborhood of modes and the heavy tails of a distribution. A BO utility is not interested in such queries. On the other extreme, in AGPR the objective is to learn the function uniformly well. This means in the same figures, AGPR will query point (x). However, given sufficient smoothness, we know that the joint probability will be very low there after exponentiation due to points (3) and (4). Therefore, the BAPE active learner will not be as interested in (x) as AGPR. Observe that the unceraity at (x) is large in the log joint probability space in comparison to the uncertainty elsewhere; however, in the probability space this is smaller than the uncertainty at the high probability regions. As Figure 5 indicates, while we model the log joint probability as a GP we are more interested in the uncertainty model of the posterior/joint probability. Finally, as a special case for BQ, [20] consider evaluating the model evidence–i.e. the integral under the conditional. Their utility function uses approximations tailored to estimating the integral well. Note that our goal of estimating the posterior well is more difficult than estimating an integral under the conditional as the former implies the latter but not vice versa.

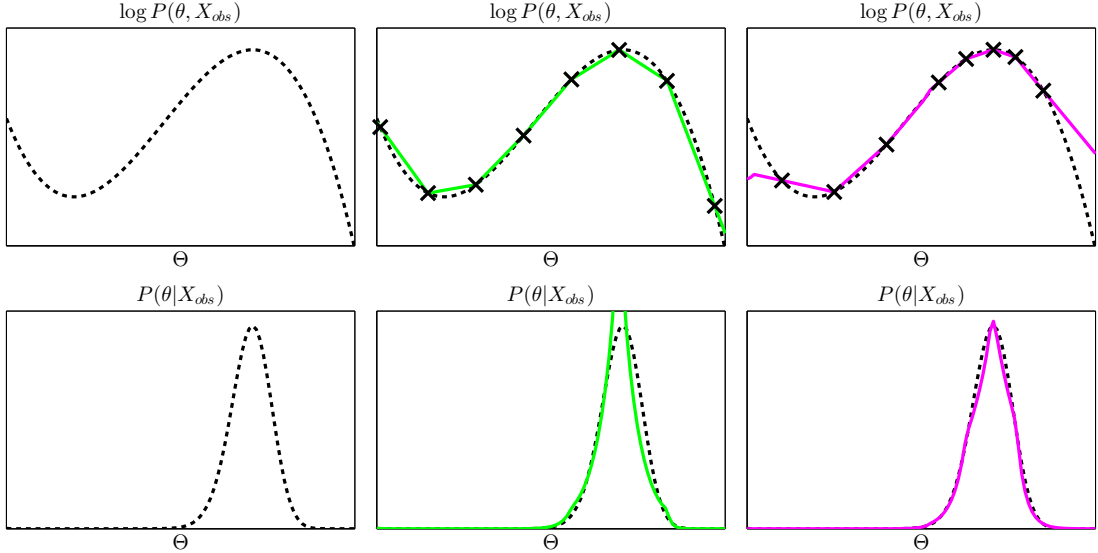

Figure 5: The first column shows the log joint probability and the corresponding posterior. In the second column we have estimates of the log joint and the posterior for uniformly spaced points. In the third column we have the same except that more points were chosen in high likelihood regions.

## 3  Other Algorithms for Comparison

We list and describe some potential alternatives for posterior estimation which we use in our empirical evaluation.

**1. MCMC - Density Estimation (MCMC-DE):** We implement MCMC with a Metropolis Hastings (MH) chain and use kernel density estimation (KDE) on the accepted points to estimate the posterior. When comparing MCMC against NED/EV we consider *the total number of queries* and not just those accepted. There are several variants of the MH proposal scheme and several tuning parameters. Comparing to all of them is nontrivial. We use MH in its basic form using a fixed Gaussian proposal distribution. Practitioners usually adjust the proposal based on the acceptance rate. Here, we chose the proposal manually by trying different values and picking the one that performed best within the queries used. Note that this comparison is advantageous to MCMC. In one experiment we test with Emcee [6], a popular package for Affine Invariant MCMC which automatically fine tunes the proposal bandwidth based on acceptance rate [6].

**2. MCMC - Regression (MCMC-R):** Here, as in MCMC-DE we use a MH Chain to generate the samples. However, this time we regress on the queries (not samples) to estimate the posterior. We include this procedure since MCMC can be viewed as a heuristic to explore the parameter space in high likelihood regions. We show that a principled query strategy outperforms this heuristic.

**3. Approximate Bayesian Computing (ABC):** There are several variants of ABC [18, 22]. We compare with a basic form given in [17]. At each iteration, we randomly sample $\theta$ from the prior and then sample an observation $\mathbf{X_{sim}}$ from the likelihood. If $d(\mathbf{X_{sim}}, \mathbf{X_{obs}}) < \epsilon$ we add $\theta$ to our collection. Here $d(\cdot, \cdot)$ is some metric on a sufficient statistic of the observation and $\epsilon > 0$ is a prespecified threshold. We perform a KDE on the collected samples to estimate the posterior. The performance of ABC depends on $\epsilon$: As for MCMC-DE we choose $\epsilon$ by experimenting with different values and choosing the value which gives the best performance within the queries used. We compare with total number of parameter values proposed and not just those retained. We compare with ABC only in experiments where it is possible to sample from the likelihood (in addition to evaluating the likelihood).

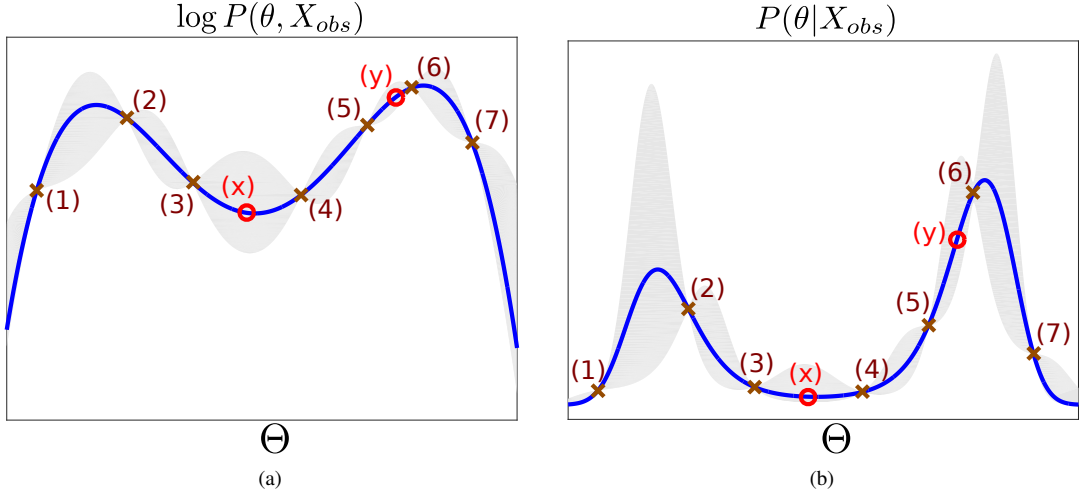

Figure 6: (a) and (b) are the true log joint probability and joint probability in blue. Assume that we have already queried at the brown crosses and let the red circles (x) and (y) be candidates for the next query. In BAPE we would be interested in querying (y) but not (x). In AGPR we would be interested in both (x) and (y) whereas in BO we would be keen in neither.

**4. Uniform Random Samples (RAND):** Here, we evaluate the likelihood at points chosen uniformly on $\Theta$ and then regress on these points.

**5. Active Gaussian Process Regression (AGPR): & 6. Bayesian Optimisation (BO-EI):** On our synthetic problems we also compare with the GP based active learning methods discussed in Section 2.5. For BO, we use the Expected Improvement [2]. We choose points using the above criterion and then regress on these points.

# 4  Experiments

We perform experiments on a series of low and high dimensional synthetic and real astrophysical experiments.

In our experiments the NED, EV utilities were maximised by evaluating them on a grid of size $10^3 - 10^8$ depending on the dimensionality and then choosing the point with the maximum value. For numerical integration in NED, we use the trapezoidal rule. Further, since the inner expectation in Equation (4) is expensive we approximate it using a one sample estimate. NED is only tested on low dimensional problems since empirical approximation and numerical integration is computationally expensive in high dimensions. In our experiments, EV slightly outperforms NED probably since the EV utility can be evaluated exactly while NED can only approximated.

We use a squared exponential kernel in all our experiments. The bandwidth for the kernel was set to be $5n^{-1/d}$ where $n$ is the total number of queries and $d$ is the dimension. This was following several kernel methods (such as kernel regression) which use a bandwidth on the order $O(n^{\frac{-c_1}{c_2+d}})$ [9]. The constant 5 was hand tuned by experimenting with a series of independent synthetic experiments. The other GP hyper-parameters, $\sigma_f^2$ and $\sigma_n^2$ were set via cross validation every 20 iterations. When we tried setting the bandwidth via cross validation too we found that it had a tendency too choose a larger than required bandwidth in the early iterations and then get stuck without decreasing. The consequence of this behaviour is that our method might not sufficiently explore the space and hence miss out on certain regions of the likelihood. Such a phenomenon has also been observed in Bayesian Optimisation and hence the bandwidth is decreased artificially as a precautionary measure against insufficient exploration [30].

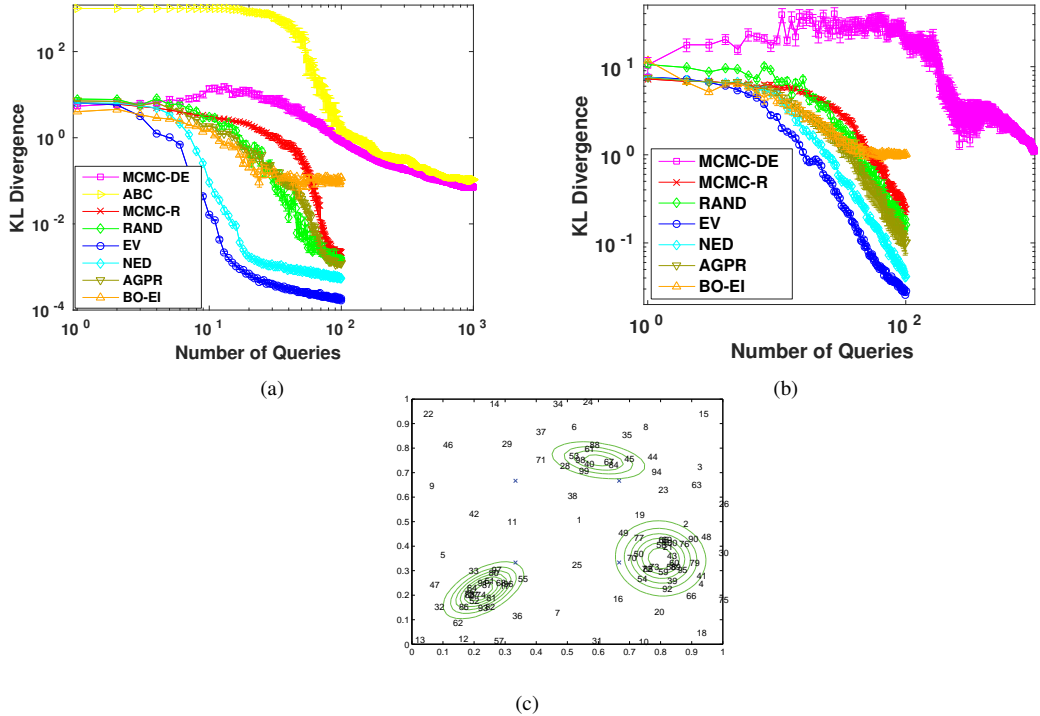

Figure 7: (a), (b): Comparison of NED/EV against MCMC-DE, ABC, MCMC-R and RAND for the 1D and 2D synthetic experiments respectively. The x-axis is the number of queries and the y-axis is the KL divergence between the truth and the estimate. All figures were obtained by averaging over 60 trials. (c): The 100 points chosen in order by NED for the 2D experiment. The green contours are the true posterior. Initially the algorithm explores the space before focusing on high probability regions.

## 4.1 Low Dimensional Synthetic Experiments

To illustrate our methods we have two simple yet instructive experiments. In the first, the parameters space is $\Theta = (0,1)$ equipped with a Beta$(1.2, 1)$ prior. We draw $\theta$ from the prior, and then draw $500$ samples from a Bernoulli $(\theta^2 + (1-\theta)^2)$ distribution: i.e. $\mathbf{X_{obs}} \in \{0,1\}^{500}$. The ambiguity on the true value of $\theta$ creates a bimodal posterior. Figure 7(a) compares NED/EV against the other methods as a function of the number of queries. For ABC, we rejected if $\left( \sum_i \mathbf{X_{obs}}^{(i)} - \sum_i \mathbf{X_{sim}}^{(i)} \right) / \sum_i \mathbf{X_{obs}}^{(i)} > 0.02$.

The second experiment is a 2D problem with $\Theta = (0,1)^2$. We artificially created a 3-modal log-joint posterior shown by green contours in Figure 7(c). Figure 7(b) compares all methods. Figure 7(c) shows the points chosen by the NED query strategy in order. We have learned the high log joint probability regions well at the expense of being uncertain at low log joint probability areas. However, this does not affect the posterior significantly as they are very small after exponentiation. ABC does not apply here since we artificially constructed the log posterior and cannot sample from the likelihood.

Our methods outperform existing methods and other heuristics by orders of magnitude on these simple experiments. Both MCMC-DE and ABC require a large number of samples before being competitive with the methods using regression. This corroborates an earlier remark that using the evaluated likelihood values in the estimate can be useful when the queries are expensive. Note that the KL divergence for BO gets stuck without decreasing further. This is because after a certain stage, most evaluations are centred near the maximum. As a consequence, the heavy tails and other modes are not explored properly.

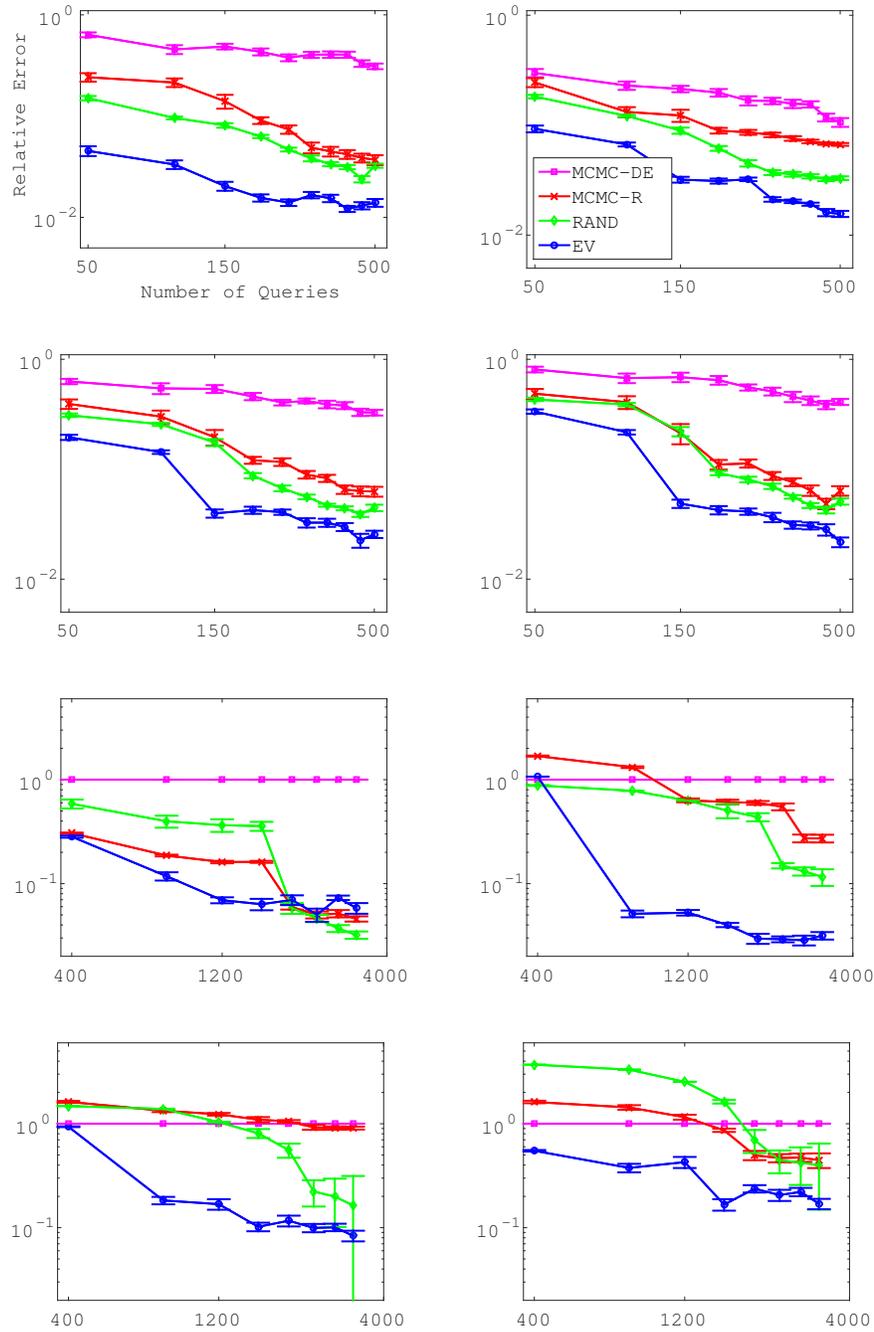

Figure 8: The first row is for the functionals $T_1, T_2$ in $d = 5$ dimensions and the second for is for the functionals $T_3, T_4$. The last two rows are the same four functionals for $d = 15$. The x-axis is the number of queries and the y-axis is $|\widehat{T}_i - T_i|/|T_i|$. We use 500 queries for $d = 5$ and 3200 queries for $d = 15$. All figures were obtained by averaging over 30 trials.

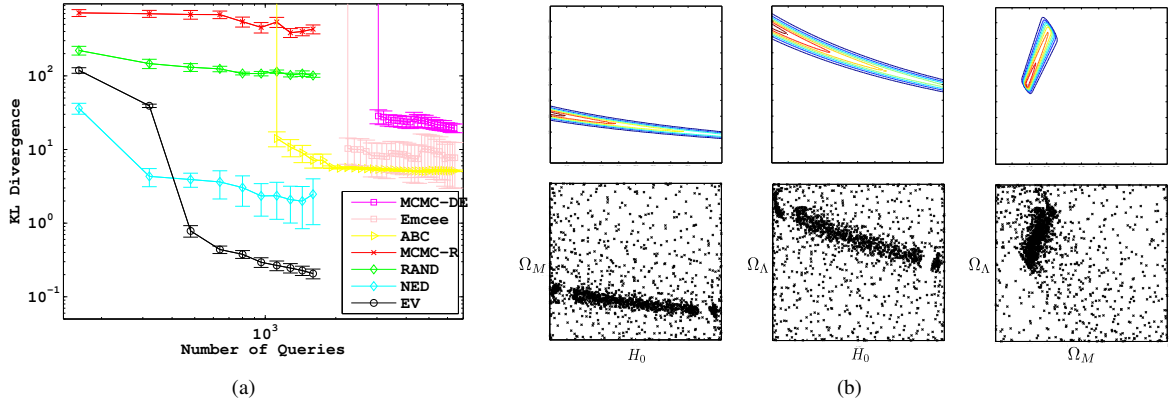

Figure 9: (a): Comparison of NED/EV against MCMC-DE, ABC, Emcee, MCMC-R and RAND on the Type Ia Supernovae dataset. For all regression methods we show results for up to 1600 queries and up to 4 times as many for MCMC and ABC. For evaluation, KL was approximated via numeric integration on a $(100)^3$ grid. Note that MCMC and ABC require several queries before a nontrivial KL with the truth is obtained. All curves were obtained by averaging over 30 runs. (b): Projections of the points selected by EV (bottom row) and the marginal distributions (top row).

## 4.2 Higher Dimensional Synthetic Experiments

We test in $d = 5$ and $15$ dimensions. We construct an artificial log likelihood so that the resulting posterior is mixture of 2 Gaussians centred at $\mathbf{0}$ and $\mathbf{1}$. Both Gaussians had covariance $\sigma^2 I_d$ where $\sigma = 0.5\sqrt{d}$. We evaluate performance by the ability to estimate certain linear functionals. The exact value of these functionals can be evaluated analytically since we know the true posterior. We use a uniform prior. Our log-likelihood, functionals and their true values are

$$\ell(\theta) = \log\left(0.5\mathcal{N}\!\left(\theta;\;\mathbf{0},\,\frac{d}{4}I_d\right) + 0.5\mathcal{N}\!\left(\theta;\;\mathbf{1},\,\frac{d}{4}I_d\right)\right)$$

$$T_1 = \mathbb{E}\left[\sum_{i=1}^{d} X_i\right] = \frac{d}{2}, \qquad T_3 = \mathbb{E}\left[\sum_{i=1}^{d-2} X_i^2 X_{i+1}\right] = \frac{d-1}{2}(1+\sigma^2),$$

$$T_2 = \mathbb{E}\left[\sum_{i=1}^{d} X_i^2\right] = \frac{d}{2}(1+2\sigma^2), \qquad T_4 = \mathbb{E}\left[\sum_{i=1}^{d-2} X_i X_{i+1} X_{i+2}\right] = \frac{d-2}{2}$$

For MCMC-DE, we draw samples $Z_1, Z_2, \ldots$ from the true likelihood. To estimate $T_i = \mathbb{E}[\phi_i(X)]$ we use the empirical estimator $\widehat{T}_i = 1/N \sum_k \phi_i(Z_k)$. Here $\phi_1 = \sum_{i=1}^{d} X_i$ for $T_1$ etc. For MCMC-DE we experimented with Gaussian proposal distributions with For EV, MCMC-R and RAND we first use the queried points to obtain an estimate of the log-likelihood by regressing on the likelihood values as explained before. Then we run an MCMC chain on this *estimate* to collect samples and use the empirical estimator for the functionals. Note that evaluating the estimate, unlike the likelihood, is cheap. We did not try NED since numerical integration is intractable in high dimensions. ABC does not apply in this experiment as we cannot sample from the log likelihood. For the proposal distributions for MCMC-DE and MCMC-R methods we used a Gaussian with standard deviations $\{0.25\sigma, 0.5\sigma, \sigma, 2\sigma, 4\sigma\}$ and report the one that performed best within the alotted queries. When applying MCMC on the regression estimates in EV, MCMC-R and RAND we used a Gaussian proposal with standard deviation $\sigma$. The results are shown in Figure 8. They demonstrate the superiority of our query strategy over the alternatives.

## 4.3 Type Ia Supernovae

We use supernovae data for inference on 3 cosmological parameters: Hubble Constant ($H_0 \in (60, 80)$), Dark Matter Fraction $\Omega_M \in (0, 1)$ and Dark Energy Fraction $\Omega_\Lambda \in (0, 1)$. The likelihood for the experiment is given by the Robertson–Walker metric which models the distance to a supernova given the parameters and the observed red-shift.

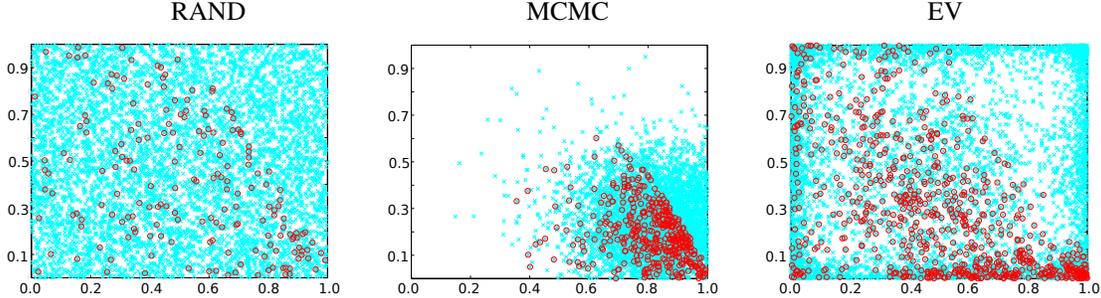

Figure 10: The projections of the first 6000 points queried by RAND MCMC, and EV respectively on to the first 2 dimensions in cyan. The points shown in red are queries at high likelihood ($\log P > -50$) points.

The dataset is taken from [5]. The parameter space is taken to be $\Theta = (0,1)^3$ (For $H_0$ we map it to $(60, 80)$ using an appropriate linear transform). We test NED/EV against MCMC-DE, ABC, MCMC-R, RAND and Emcee, a popular python package for affine invariant MCMC. For ABC, sampling from the likelihood is as expensive as computing the likelihood. Figure 9(a) compares all methods. Figure 9(b) shows the points queried by EV and the marginals of the true posterior. As expected, most of EV's queries are concentrated around the modes and heavy tails of the posterior. The KL for RAND decreases slowly since it accumulates points at the high likelihood region very slowly. MCMC-R performs poorly since it has only explored part of the high likelihood region. For NED/EV after an initial exploration phase, the error shoots down.

The likelihood evaluations in this experiment are quite cheap, taking only a fraction of a second for each query. Determining the next point is more expensive in EV than methods such as MCMC, ABC and RAND due to matrix inversion in GPs. However, we found that the latter methods required up to 20-30 times the number of likelihood evaluations to be competitive with EV in this experiment. Therefore, despite the fact that the EV query strategy is expensive it performs better than other methods on wall clock time. This illustrates that principled adaptive query strategies can reap great dividends in posterior estimation.

## 4.4 Luminous Red Galaxies

Here we used data on Luminous Red Galaxies (LRGs) for inference on 8 cosmological parameters: spatial curvature $\Omega_k \in (-1, 0.9)$, dark energy fraction $\Omega_\Lambda \in (0, 1)$, cold dark matter density $\omega_c \in (0, 1.2)$, baryonic density $\omega_B \in (0.001, 0.25)$, scalar spectral index $n_s \in (0.5, 1.7)$, scalar fluctuation amplitude $A_s \in (0.65, 0.75)$, running of spectral index $\alpha \in (-0.1, 0.1)$ and galaxy bias $b \in (0, 3)$. The likelihood is obtained via the Galaxy Power spectrum which measures the distribution of temperature fluctuations as a function of scale. We use software and data from [29]. Our parameter space is taken to be $(0,1)^8$ by appropriately linear transforming the range of the variables. Each query takes about 4-5 seconds. In EV, determining the next point takes about 0.5-1 seconds with $\approx 2000$ points and about 10-15 seconds with $\approx 10000$ points. In this regime, where the cost of the likelihood evaluation is more expensive or comparable to the cost of determining the next point in EV, we significantly outperforms other methods on wall clock time. We do not compare with NED due to the difficulty of high dimensional numerical integration. We do not compare with ABC since the software only permits evaluation of the likelihood but not sampling.

Figure 10 shows points queried by MCMC, RAND and EV projected on the first 2 dimensions. MCMC has several high likelihood points but its queries are focused on a small region of the space. RAND does not have many points at high likelihood regions. EV has explored the space fairly well and at the same time has several queries at high likelihood regions. As numerical integration in 8 dimensions is difficult, we cannot obtain ground truth for this experiment. Therefore, we perform the following simple test. We queried $250,000$ points uniformly at random from the parameter space to form a test set. We then run EV, MCMC-R and RAND for up to $12,000$ queries to collect points and estimate the posterior. Performance is evaluated by the mean squared reconstruction error of the *exponentiated* log joint probabilities (joint probabilities). Figure 11 shows the results. The error for RAND and MCMC-R stay the same throughout since the problem is difficult and they did not have sufficient number of high likelihood points throughout the space.

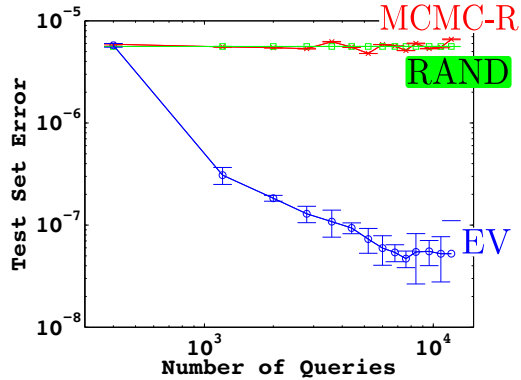

Figure 11: Comparison of EV against MCMC-R and RAND. We use up to 12000 queries for all methods. The y-axis is the mean squared reconstruction error. The curves were obtained by averaging over 16 runs.

## 5 Conclusions

We proposed a framework for query efficient posterior estimation for expensive blackbox likelihood evaluations. Our methods use GPs and are based on popular ideas in Bayesian active learning. We demonstrate that our methods outperform natural alternatives in practice.

Note that in Machine Learning it is uncommon to treat posterior estimation in a regression setting. This is probably since the estimate will depend on the intricacies of the regression algorithm. Thus if likelihood evaluations are inexpensive, MCMC seems like a natural choice due to its theoretical guarantees in the large sample regime. However, our work demonstrates that when likelihood evaluations are expensive, such as in scientific simulations, treating posterior estimation in an active regression framework enables us to be significantly query efficient.

The proposed methods do not scale very well with dimension, which is a common problem with nonparametric methods. Going forward we wish to tackle active posterior estimation in several dozens of dimensions. A possible avenue would be to use ideas from some recent progress on high dimensional Bayesian Optimisation using additive models [11].

## Acknowledgement

This research was partly funded by DOE grant DESC0011114. An abridged version of this work appeared in Kandasamy et al. [10].